\documentclass{article}

\PassOptionsToPackage{numbers, compress}{natbib}

\usepackage[preprint]{neurips_2023}

\usepackage{longtable}
\usepackage[utf8]{inputenc} 
\usepackage[T1]{fontenc}   
\usepackage{hyperref}      
\usepackage{url}            
\usepackage{booktabs}       
\usepackage{amsfonts}       
\usepackage{CJKutf8}
\usepackage{nicefrac}       
\usepackage{microtype}      
\usepackage{xcolor}        
\usepackage{amsmath}
\usepackage{tcolorbox}
\usepackage{makecell} 
\usepackage{array}    
\usepackage{float}
\usepackage{graphicx}
\usepackage{ltablex}
\usepackage{lipsum}
\usepackage{graphicx}
\usepackage{wrapfig}
\usepackage[size=small]{caption}
\usepackage{mathtools}
\usepackage{tabularx}
\usepackage{amssymb}
\usepackage{amsthm}
\usepackage{soul}

\definecolor{textblue}{rgb}{.2,.2,.7}
\definecolor{textred}{rgb}{0.54,0,0}
\definecolor{textblack}{rgb}{0,0,0}
\definecolor{textgreen}{rgb}{0,0.53,0}
\usepackage{listings}
\lstdefinestyle{pythonstyle}{
    language=Python,
    morekeywords={None}, 
    breaklines=true,
}

\usepackage{array}
\newcolumntype{M}[1]{>{\centering\arraybackslash}m{#1}}

\title{Octopus v3: Technical Report for On-device Sub-billion Multimodal AI Agent}

\author{Wei Chen \thanks{Corresponding author} \\
Stanford University\\
\texttt{\{weichen6\}@stanford.edu} \\
\And
Zhiyuan Li \\
Stanford University\\
\texttt{\{zhiyuan8\}@stanford.edu} \\
}

\begin{document}
\begin{CJK*}{UTF8}{gbsn}
\nocite{*}
\maketitle

\begin{abstract}
A multimodal AI agent is characterized by its ability to process and learn from various types of data, including natural language, visual, and audio inputs, in order to inform its actions. Despite advances in large language models that incorporate visual data, such as GPT-4V, effectively translating image-based information into actionable outcomes for AI agents remains challenging. In this paper, we introduce a multimodal model that incorporates the concept of the \textit{functional token}, designed specifically for AI agent applications. To ensure compatibility with edge devices, our model is optimized to a compact size of fewer than \textbf{1B} parameters. Like GPT-4, our model can process both English and Chinese. We demonstrate that the model operates efficiently on a wide range of edge devices, including ones as constrained as a Raspberry Pi. Demo videos are available at this \href{https://octopus3.nexa4ai.com}{link}. Access to the model weights and inference code can be requested through this \href{https://www.nexa4ai.com/apply}{link}; the model is still under testing and is currently intended for research purposes only. 
\end{abstract}

\begin{figure}[h]
    \centering
    \includegraphics[width=1.0\textwidth]{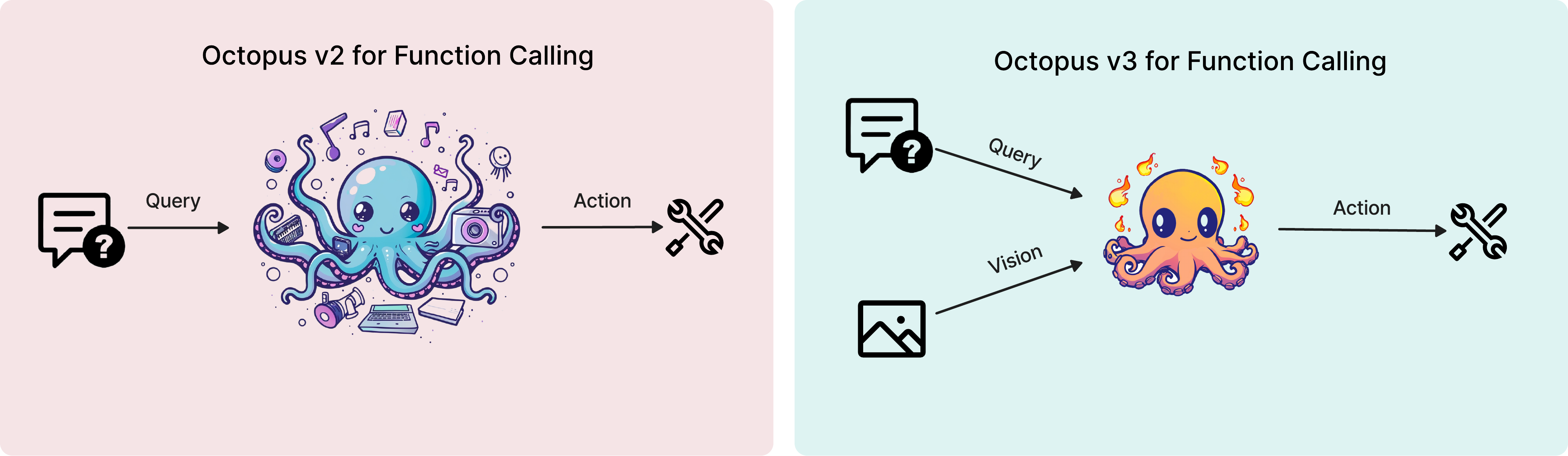}
    \caption{A sub-billion-parameter multimodal AI agent with vision capabilities}
    \label{fig:vision-agent}
\end{figure}

\section{Introduction}
The rapid advancement of artificial intelligence has revolutionized the way we interact with technology, giving rise to the development of sophisticated AI agents capable of performing complex tasks and making decisions based on various forms of input, including natural language \cite{xi2023rise,srinivasan2023nexusraven,wang2024mobile,dong2023towards,hauptman2023adapt} and visual information \cite{chattopadhyay2017evaluating,kolve2017ai2}. These agents have the potential to automate a wide range of processes, from simple tasks like image recognition and language translation to more complex endeavors such as medical diagnosis \cite{gilbert2023large,peng2023study} and autonomous navigation \cite{cui2024survey,fu2024drive}. At the heart of these AI agents are multimodal language models, which enable them to understand and generate human-like responses by processing and integrating multiple modalities of data, such as text, images, and even audio or video.

Multimodal language models represent a significant leap forward from traditional language models, which primarily focus on processing and generating text. By incorporating visual information, these models can better understand the context and semantics of the input data, leading to more accurate and relevant outputs. For example, a multimodal language model trained on a dataset of images and their corresponding captions would be able to generate more descriptive and contextually appropriate captions for new images compared to a text-only model. The ability to process and integrate multiple modalities of data is crucial for the development of multimodal AI agents, as it allows them to perform tasks that require an understanding of both language and visual information, such as visual question answering, image-guided navigation, and multimodal sentiment analysis.

One of the key challenges in developing multimodal language models is effectively encoding visual information into a format that can be processed by the model. This is typically achieved through the use of convolutional neural networks (CNNs) \cite{li2021survey} or transformer-based architectures \cite{vaswani2017attention}, such as the Vision Transformer (ViT) \cite{arnab2021vivit}. CNNs have been widely used in computer vision tasks for their ability to extract hierarchical features from images, allowing the model to learn increasingly complex representations of the input data. On the other hand, transformer-based architectures like ViT have recently gained popularity due to their ability to capture long-range dependencies and model global context, which is particularly useful for understanding the relationships between objects in an image. These architectures enable the model to extract meaningful features from the input image and convert them into a vector representation that can be integrated with the text input.

Another approach to encoding visual information is through image tokenization, which involves breaking down the image into smaller, discrete units or tokens. This method allows the model to process the image in a manner similar to how it processes text, enabling a more seamless integration of the two modalities. Tokenized image information can be fed into the model alongside the text input, allowing the model to attend to both modalities simultaneously and generate more accurate and contextually relevant outputs. For instance, the DALL-E model developed by OpenAI uses a variant of the VQ-VAE (Vector Quantized Variational Autoencoder) \cite{razavi2019generating,peng2021generating} to tokenize images, enabling the model to generate novel images based on textual descriptions.

The development of smaller, more efficient models that are capable of taking action based on a user's query and a provided image has significant implications for the future of AI agents. These models can be deployed on resource-constrained devices, such as smartphones or IoT devices, enabling a wider range of applications and use cases. By leveraging the power of multimodal language models, these smaller agents can understand and respond to user queries in a more natural and intuitive manner, while also considering the visual context provided by the user. This opens up possibilities for more engaging and personalized interactions, such as virtual assistants that can provide visual recommendations based on a user's preferences or smart home devices that can adjust settings based on the user's facial expressions.

Moreover, the development of multimodal AI agents has the potential to democratize AI technology, making it more accessible to a broader range of users and industries. Smaller, more efficient models can be trained on less powerful hardware, reducing the computational resources and energy consumption required for deployment. This could lead to the widespread adoption of AI agents in various domains, from healthcare and education to entertainment and e-commerce, ultimately transforming the way we live and work.

\section{Related work}

\textbf{Multimodal models} \quad Multimodal models have gained significant attention due to their ability to process and learn from various data types, such as text, images, and audio \cite{liu2024visual,liu2023improved}. These models capture the complex interactions between different modalities and leverage their complementary information to improve performance on a wide range of tasks. Vision-Language Pre-training (VLP) models, such as ViLBERT \cite{lu2019vilbert}, LXMERT \cite{tan2019lxmert}, and VisualBERT \cite{li2019visualbert}, learn to align visual and textual features through cross-modal attention, generating rich multimodal representations. Multimodal transformer architectures, like the Multimodal Transformer (MMT) \cite{tsai2019multimodal} and the Vision-and-Language Transformer (ViLT) \cite{kim2021vilt}, adapt the transformer architecture to handle multiple modalities efficiently. Researchers have also explored incorporating other modalities, such as audio and facial expressions, into models like the Multimodal Sentiment Analysis (MSA) model \cite{soleymani2017survey} and the Multimodal Emotion Recognition (MER) \cite{sebe2005multimodal} model. By leveraging complementary information from different modalities, multimodal models achieve better performance and generalization compared to unimodal approaches.


\textbf{On-device language model development} \quad In this paper, we define on-device language models as those with fewer than 7 billion parameters, since we find it very difficult to run a 13B model on an edge device even with quantization. Recent efforts in this area include Google's Gemma 2B and 7B \cite{gemma-2023-open-models}, Stability AI's Stable Code 3B \cite{stable-code-3b}, and Meta's Llama 7B \cite{touvron2023llama}. Interestingly, research from Meta suggests that smaller language models perform better with deep and thin architectures, unlike their larger counterparts. Other beneficial techniques for on-device models include embedding sharing, grouped-query attention, and immediate block-wise weight sharing, as introduced in the MobileLLM \cite{liu2024mobilellm} paper. These findings highlight the importance of considering alternative design methods when developing smaller language models for on-device applications, as they may require different optimization approaches compared to larger models.

\section{Methodology}
This section discusses the primary techniques employed in the development of the Octopus v3 model. The two critical aspects of multimodal model development are integrating image information with text input and optimizing the model's ability to predict actions.

\subsection{Encoding visual information}
Numerous methods exist for encoding visual information in image processing, with embeddings from hidden layers being commonly utilized. For example, the VGG-16 \cite{tammina2019transfer} model's hidden layer embeddings are used in style transfer tasks. OpenAI's CLIP \cite{radford2021learning} model demonstrates the ability to align embeddings from both text and images, leveraging its image encoder for embedding images. More advanced techniques, such as image tokenization, are employed in approaches like ViT \cite{liu2024visual}.

Our research evaluated various image encoding techniques and found the CLIP model's method to be the most effective. Consequently, the application presented in this paper utilizes a CLIP-based model for image encoding.

\subsection{Functional token}

Analogous to tokenization applied to natural language and images, particular functions can be encapsulated as \textit{functional tokens} \cite{chen2024octopus2,chen2024octopus}. We introduce a training strategy for these tokens that mirrors the techniques used by natural language models to handle unseen terms. This methodology is akin to the word2vec scheme \cite{mikolov2013efficient}, in which the meaning of a token is enriched by its surrounding context. For instance, advanced language models might initially struggle with complex chemical terms such as \textit{PEGylation} and \textit{endosomal escape}. Nevertheless, these models are capable of acquiring such terminology through causal language modeling, especially when trained on datasets containing these terms. Similarly, functional tokens can be learned using parallel strategies, with the Octopus v2 model serving as a robust platform for such learning processes. Our research shows that the space of definable functional tokens is effectively unlimited, allowing any particular function to be represented as a token.

\subsection{Multi-stage training}

To develop a high-performing multimodal AI agent, we employ a model architecture that integrates a causal language model with an image encoder. The training process for this model is structured in multiple stages. Initially, the causal language model and the image encoder are trained separately to establish a foundational baseline model. Subsequently, these components are merged, and the model undergoes alignment training to synchronize image and text processing capabilities. Following this, the training incorporates methodologies from the Octopus v2 \cite{chen2024octopus2} framework to facilitate the learning of \textit{functional tokens}. In the final stage of training, these functional tokens, which are capable of interacting with the environment, provide feedback that is used to further refine and optimize the model. Specifically, in this final stage we apply reinforcement learning, employing another large language model as the reward model. This iterative training approach enhances the model's ability to process and integrate multimodal information effectively.

\section{Model evaluation}
In this section, we present the results of our model and compare them with the outcomes achieved by combining the GPT-4V and GPT-4 models \cite{achiam2023gpt,brown2020language,radford2019language,radford2018improving}. For this comparison, we first employ GPT-4V (\textit{gpt-4-turbo}) to process the image information. The extracted information is then used within the GPT-4 framework (\textit{gpt-4-turbo-preview}), incorporating all functional descriptions into the context and applying few-shot learning to enhance performance. In our demonstration, we convert 10 commonly used smartphone APIs into functional tokens and evaluate their performance, as detailed in the subsequent sections.

It is important to note that although only 10 functional tokens are showcased in this paper, the model can be trained with a broader array of tokens to create a more versatile AI agent. Our findings indicate that our model, with fewer than \textbf{1B} parameters, performs comparably to the combination of GPT-4V and GPT-4 as a multimodal AI agent for the selected APIs.

Furthermore, the scalability of our model allows for the inclusion of a wide range of functional tokens, enabling the creation of highly specialized AI agents tailored to specific domains or use cases. This adaptability makes our approach particularly valuable in industries such as healthcare, finance, and customer service, where AI-driven solutions can significantly enhance efficiency and user experience.

For all the function names below, the Octopus model outputs only the functional tokens, such as \texttt{<nexa\_0>,...,<nexa\_N>}; we replace each functional token with the corresponding function name for clearer presentation. All the results below are direct generations that do not require any output parser. Octopus v3 is a single model that can handle both Chinese and English, meaning that training a separate model specifically for Chinese is unnecessary.

\subsection{Send email}

\begin{table}[H]
\centering
\begin{tabular}{@{}M{3cm}M{9cm}@{}}
 \toprule
 Text and vision input & \vspace{1mm} \includegraphics[width=6cm]{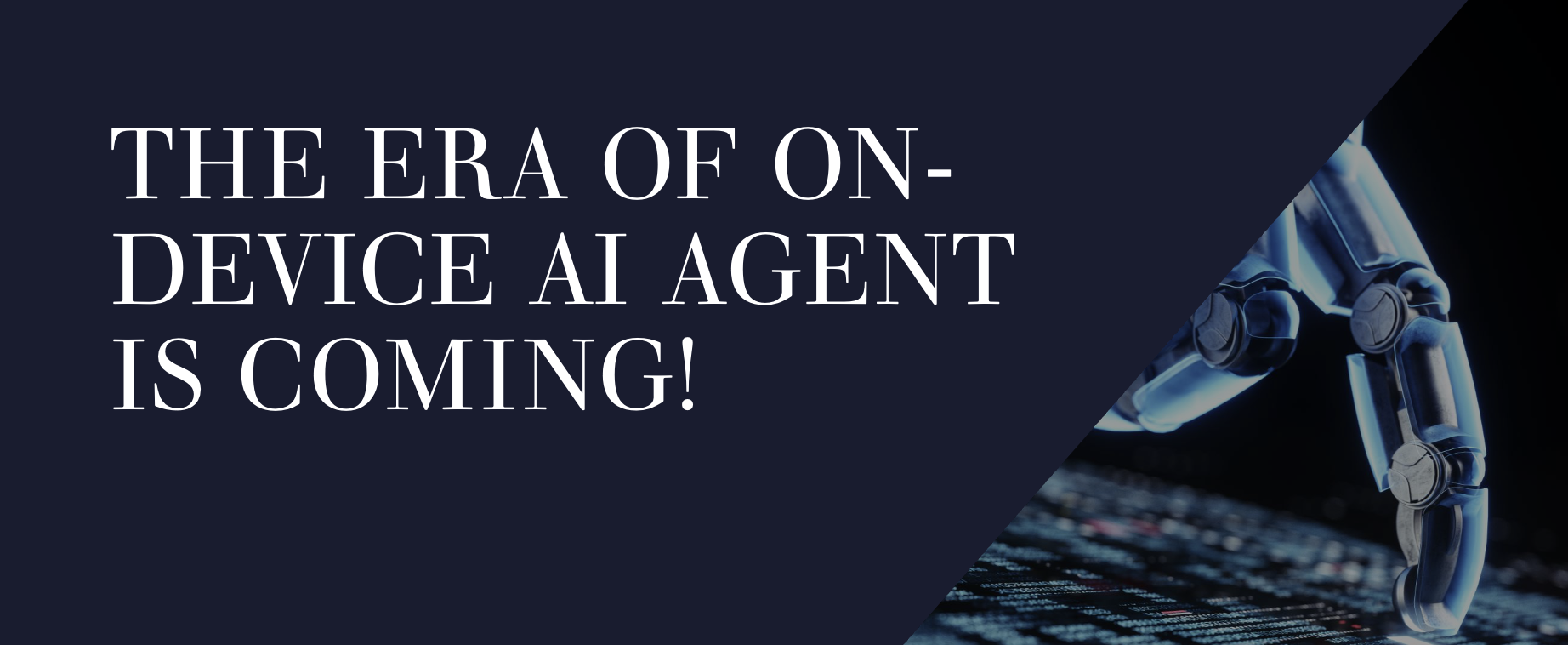} \vspace{1mm}\\ 
 & send an email to hello@nexa4ai.com with the title as AI progress. Be short, and capture the main idea;告诉hello@nexa4ai.com, 标题是AI进展 \\ 
 \midrule
 Octopus v3 & \texttt{send\_email('hello@nexa4ai.com', 'AI progress', 'The era of on device AI agent is coming! With the latest technology, businesses and individuals are embracing the power of technology in various aspects of life. The focus on innovation and efficiency in the current era is unparalleled.')<nexa\_end>} \\
 \addlinespace
 Octopus v3 Chinese & \texttt{send\_email('hello@nexa4ai.com', 'AI进展', 'The era of on device AI agent is coming!')<nexa\_end>} \\
 \addlinespace
  GPT-4V + GPT-4 & \texttt{send\_email('hello@nexa4ai.com', 'AI progress', 'The era of on-device AI agents is coming! Advancements in AI technology are leading to more powerful and autonomous on-device AI solutions.')}    \\
 \bottomrule
\end{tabular}
\caption{Send email example}
\end{table}

\subsection{Send text message}
\begin{table}[H]
\centering
\begin{tabular}{@{}M{3cm}M{9cm}@{}}
 \toprule
 Text and vision input & \vspace{1mm} \includegraphics[width=5cm]{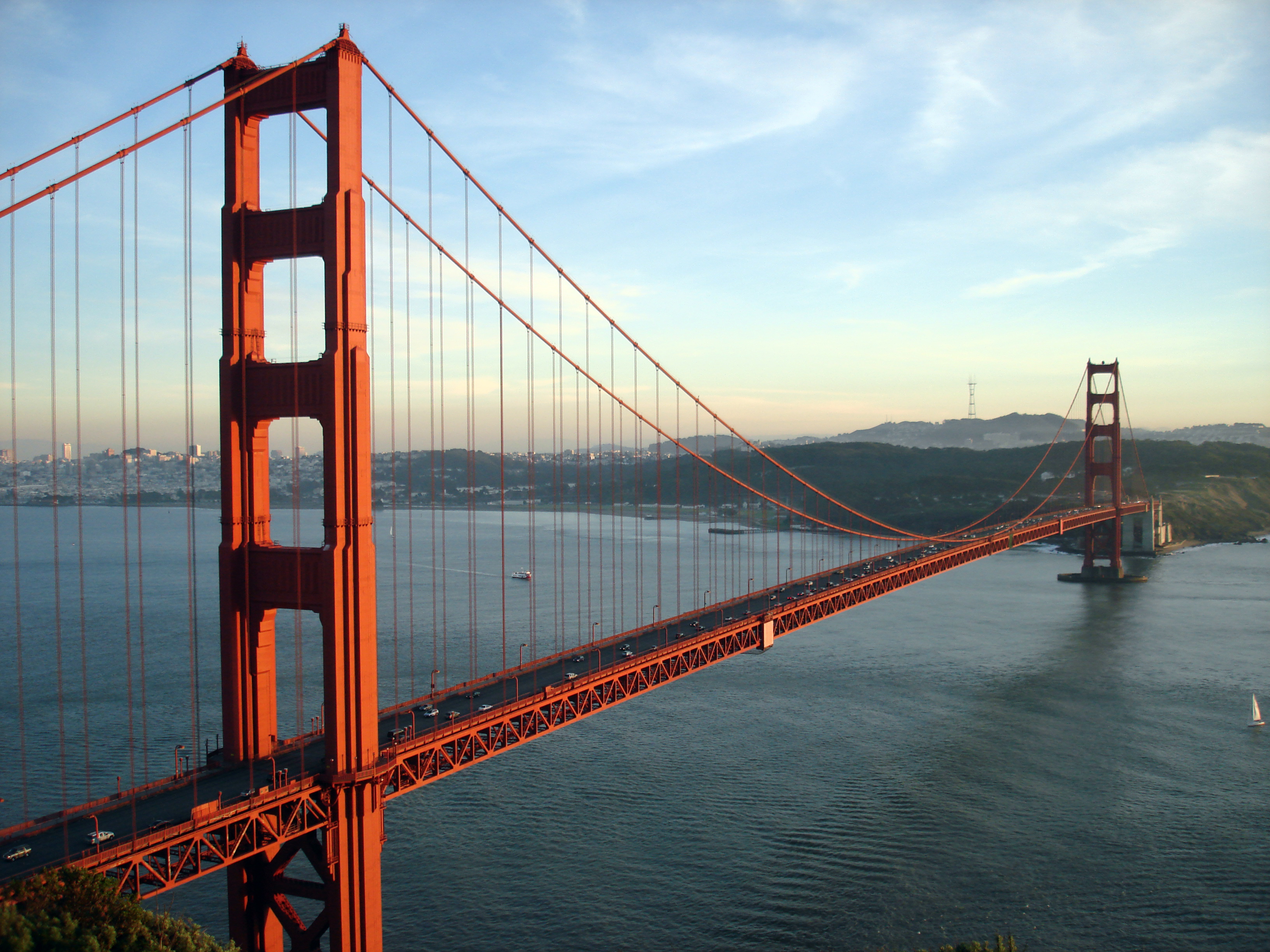} \vspace{1mm}\\ 
 & send text message to 1234567890； 发个text message给1234567890 \\ 
 \midrule
 Octopus v3 & \texttt{send\_text\_message('Golden bridge with cityscape in background, under a serene sky.', '1234567890')<nexa\_end>} \\
 \addlinespace
  Octopus v3 Chinese& \texttt{send\_text\_message('Golden bridge with cityscape in the background, early evening.', '1234567890')<nexa\_end>} \\
 \addlinespace
  GPT-4V + GPT-4 & \texttt{send\_text\_message('Picture of the Golden Gate Bridge in San Francisco, taken at dusk with clear skies and the city skyline in the background.', "1234567890")}
    \\
 \bottomrule
\end{tabular}
\caption{Send text message example}
\end{table}

\subsection{Google search example}
\begin{table}[H]
\centering
\begin{tabular}{@{}M{3cm}M{9cm}@{}}
 \toprule
 Text and vision input & \vspace{1mm} \includegraphics[width=8cm]{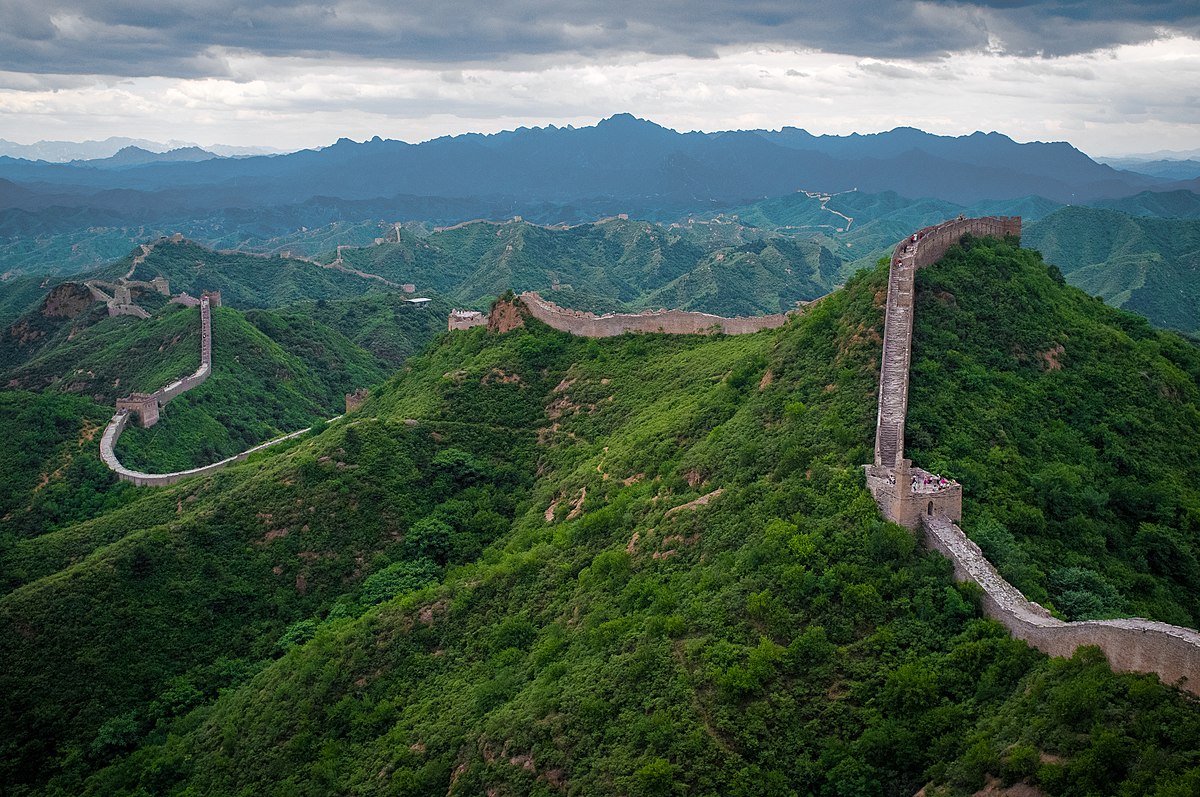} \vspace{1mm}\\ 
 & Tell me the history of this scene; 帮助我搜索它的历史 \\ 
 \midrule
 Octopus v3 & \texttt{google\_search('history of the Chinese wall on the mountains')<nexa\_end>} \\
 \addlinespace
  Octopus v3 Chinese& \texttt{google\_search('history of the Great Wall of China')<nexa\_end>} \\
 \addlinespace
  GPT-4V + GPT-4 & \texttt{google\_search('Great Wall of China history')}    \\
 \bottomrule
\end{tabular}
\caption{Google search example}
\end{table}

\subsection{Amazon purchase example}
\begin{table}[H]
\centering
\begin{tabular}{@{}M{3cm}M{9cm}@{}}
 \toprule
 Text and vision input & \vspace{1mm} \includegraphics[width=6cm]{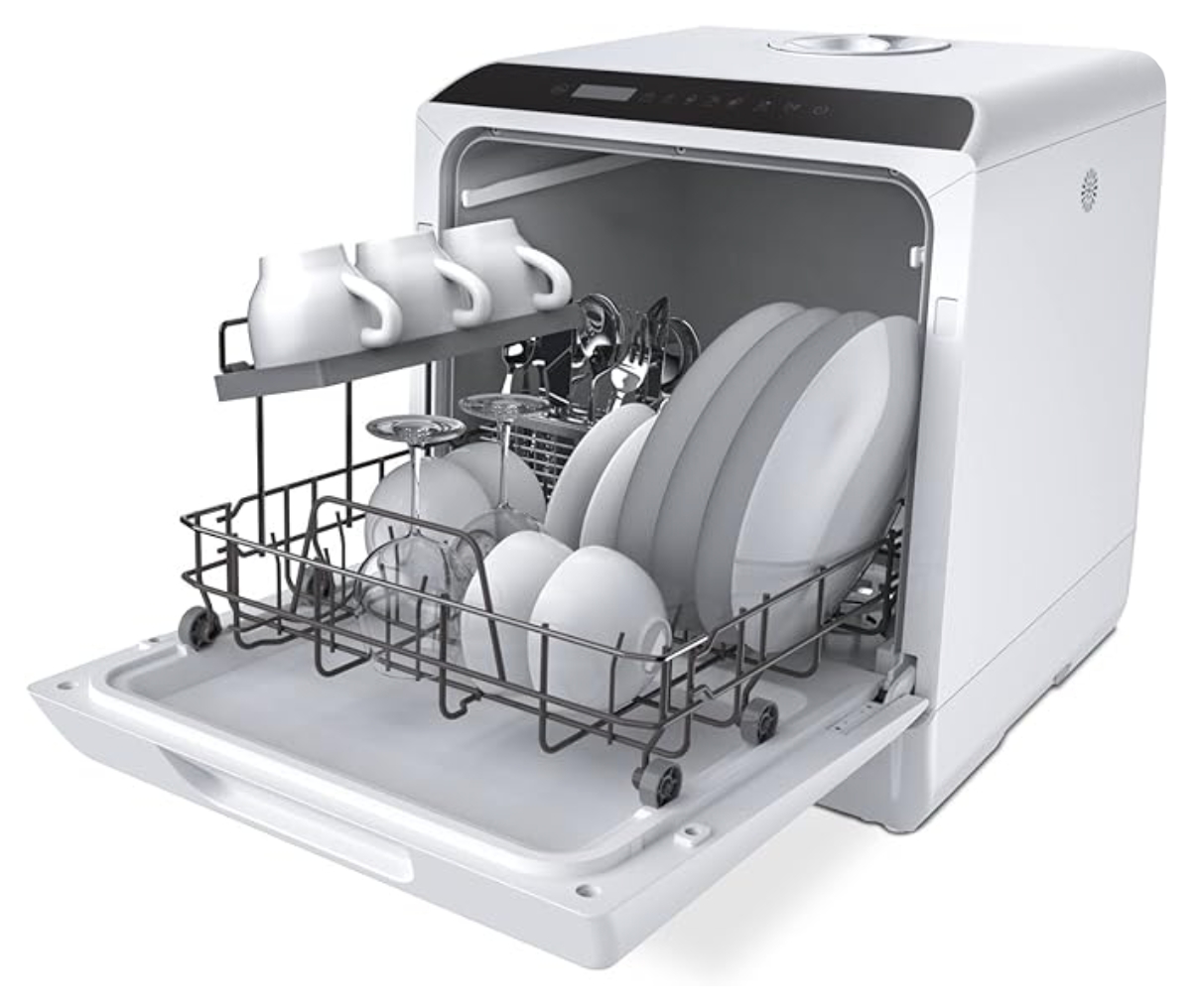} \vspace{1mm}\\ 
 & I want to purchase this; 我要买这个 \\ 
 \midrule
 Octopus v3 & \texttt{amazon\_purchase('White, compact, wall-mounted kitchen appliance for drying dishes, including white rice and black edges.', 'Kitchen \& Dining')<nexa\_end>} \\
 \addlinespace
 Octopus v3 Chinese & \texttt{amazon\_purchase('White, compact, wall-mounted kitchen appliance for drying and cleaning, including multiple bowls and a stack of plates.', 'Kitchen \& Dining')<nexa\_end>} \\
 \addlinespace
  GPT-4V + GPT-4 & \texttt{amazon\_purchase('compact countertop dishwasher', 'home appliances')}    \\
 \bottomrule
\end{tabular}
\caption{Amazon purchase example}
\end{table}

\subsection{Smart recycle}
\begin{table}[H]
\centering
\begin{tabular}{@{}M{3cm}M{9cm}@{}}
 \toprule
 Text and vision input & \vspace{1mm} \includegraphics[width=6cm]{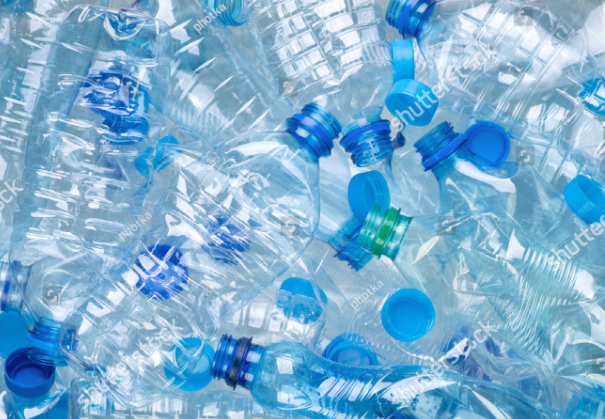} \vspace{1mm}\\ 
 & Dispose this; 回收这个\\ 
 \midrule
 Octopus v3 & \texttt{smart\_recycle('plastic water bottles', 'plastic')<nexa\_end>} \\
 \addlinespace
 Octopus v3 Chinese& \texttt{smart\_recycle('plastic water bottles', 'plastic')<nexa\_end>} \\
 \addlinespace
  GPT-4V + GPT-4 & \texttt{smart\_recycle('water bottles', 'plastic')}    \\
 \bottomrule
\end{tabular}
\caption{Smart recycle example}
\end{table}

\subsection{Lost and found}
\begin{table}[H]
\centering
\begin{tabular}{@{}M{3cm}M{9cm}@{}}
 \toprule
 Text and vision input & \vspace{1mm} \includegraphics[width=4cm]{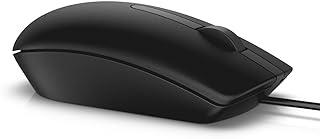} \vspace{1mm}\\ 
 & Report my loss; 报失 \\ 
 \midrule
 Octopus v3 & \texttt{lost\_and\_found('black optical computer mouse', 'Modern, black optical mouse with a streamlined design, featuring left and right click buttons and a scroll wheel, ergonomic for everyday use.')<nexa\_end>} \\
 \addlinespace
  Octopus v3 Chinese& \texttt{lost\_and\_found('black optical computer mouse', 'Modern, black optical mouse with a streamlined design, featuring left and right click buttons and a scroll wheel, ergonomic for everyday use.')<nexa\_end>} \\
 \addlinespace
  GPT-4V + GPT-4 & \texttt{lost\_and\_found('optical mouse', 'A black, wired, optical mouse. It has a standard design with two buttons and a scroll wheel.')}    \\
 \bottomrule
\end{tabular}
\caption{Lost and found example}
\end{table}

\subsection{Interior design}
\begin{table}[H]
\centering
\begin{tabular}{@{}M{3cm}M{9cm}@{}}
 \toprule
 Text and vision input & \vspace{1mm} \includegraphics[width=3cm]{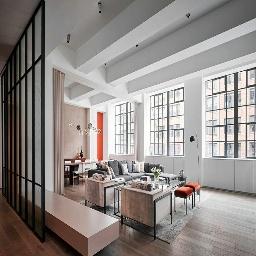} \vspace{1mm}\\ 
 & Design it; 设计这个房子 \\ 
 \midrule
 Octopus v3 & \texttt{interior\_design('living room', 'modern minimalist style with neutral tones, wood furniture, and an orange ottoman for a warm touch')<nexa\_end>} \\
 \addlinespace
  Octopus v3 Chinese& \texttt{interior\_design('living room', 'modern living room with white walls, black wooden beams, and an open-plan layout featuring a gray sectional sofa, black furniture, and an orange ottoman')<nexa\_end>} \\
 \addlinespace
  GPT-4V + GPT-4 & \texttt{interior\_design("living room", "Elegant and spacious living room with large windows, high ceilings, and a mixture of modern and mid-century furniture. Include a color scheme that complements the natural light and suggestions for decorative accessories that enhance the open, airy feel of the space.")}    \\
 \bottomrule
\end{tabular}
\caption{Interior design example}
\end{table}

\subsection{Instacart shopping}
\begin{table}[H]
\centering
\begin{tabular}{@{}M{3cm}M{9cm}@{}}
 \toprule
 Text and vision input & \vspace{1mm} \includegraphics[width=5cm]{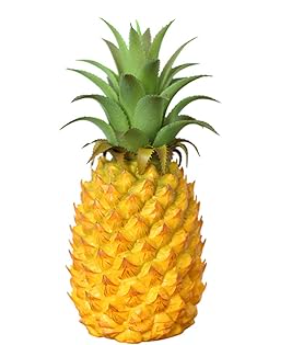} \vspace{1mm}\\ 
 & Order two for me; 帮我买两 \\ 
 \midrule
 Octopus v3 & \texttt{instacart\_shopping('pineapple', 2)<nexa\_end>} \\
 \addlinespace
  Octopus v3 Chinese& \texttt{instacart\_shopping('pineapple', 2)<nexa\_end>} \\
 \addlinespace
  GPT-4V + GPT-4 & \texttt{instacart\_shopping('pineapple', 2)}  \\
 \bottomrule
\end{tabular}
\caption{Instacart shopping example}
\end{table}

\subsection{DoorDash example}
\begin{table}[H]
\centering
\begin{tabular}{@{}M{3cm}M{9cm}@{}}
 \toprule
 Text and vision input & \vspace{1mm} \includegraphics[width=4cm]{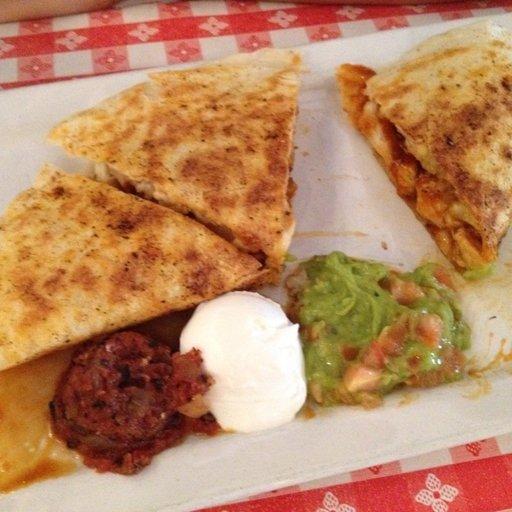} \vspace{1mm}\\ 
 &  Buy and deliver to Stanford Y2E2 building; 买了之后送到同济大学 \\ 
 \midrule
 Octopus v3 & \texttt{doordash\_order('Plate of Mexican cuisine with quesadilla, guacamole, sour cream, and spicy chorizo sausage.', ' Stanford University Y2E2')<nexa\_end>} \\
 \addlinespace
 Octopus v3 Chinese& \texttt{doordash\_order('Plate of Mexican cuisine with quesadilla, guacamole, sour cream, and spicy chorizo sausage.', '同济大学')<nexa\_end>} \\
 \addlinespace
  GPT-4V + GPT-4 & \texttt{doordash\_order('quesadilla with guacamole, sour cream, and refried beans', 'Stanford Y2E2')}    \\
 \bottomrule
\end{tabular}
\caption{DoorDash example}
\end{table}

\subsection{Animal care example}
\begin{table}[H]
\centering
\begin{tabular}{@{}M{3cm}M{9cm}@{}}
 \toprule
  Text and vision input & \vspace{1mm} \includegraphics[width=8cm]{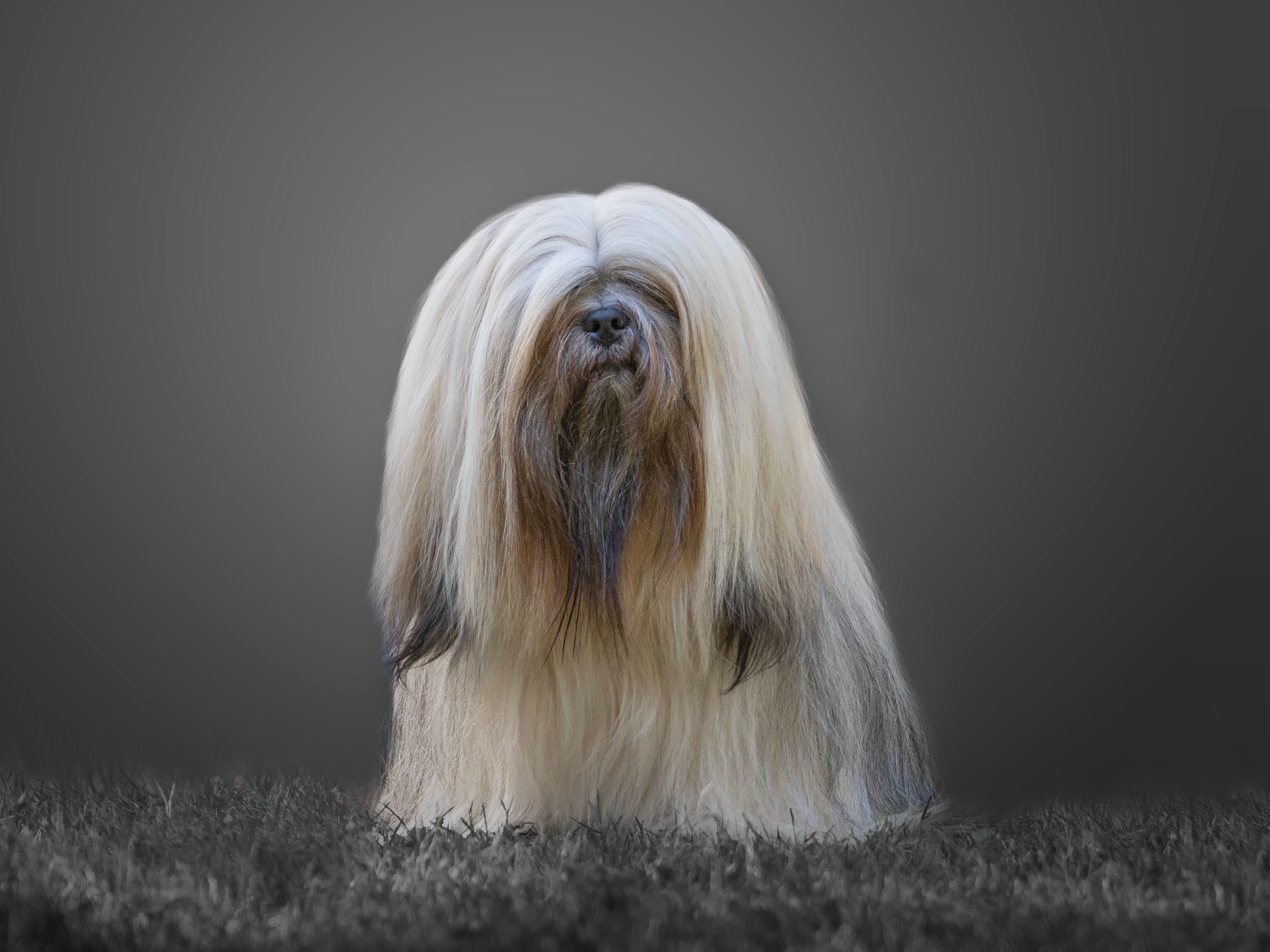} \vspace{1mm}\\ 
 & Help me take care of him; 帮我照顾他 \\ 
 \midrule
 Octopus v3 & \texttt{animal\_care('dog', 'grooming')<nexa\_end>} \\
 \addlinespace
  Octopus v3 Chinese& \texttt{animal\_care('dog', 'grooming')<nexa\_end>} \\
 \addlinespace
  GPT-4V + GPT-4 & \texttt{animal\_care('dog', 'grooming')}    \\
 \bottomrule
\end{tabular}
\caption{Animal care example}
\end{table}

\section{Social impact}
Building on the capabilities of Octopus v2, the updated model incorporates both textual and visual information, advancing beyond the purely text-based approach of its predecessor. This significant development enables the simultaneous processing of visual and natural language data, paving the way for broader applications. The \textit{functional token} introduced in Octopus v2 is adaptable and finds applications across diverse sectors, such as the healthcare and automotive industries. With the addition of visual data, the potential of the \textit{functional token} extends to areas such as \textbf{autonomous driving} and \textbf{robotics}. Furthermore, our multimodal model facilitates the practical conversion of a Raspberry Pi into devices like the Rabbit R1 \cite{rabbit2024} or the Humane AI Pin \cite{humane2024}, utilizing an on-device model instead of a cloud-based solution.

The \textit{functional token} is now licensed, and we encourage developers to engage with our framework to innovate freely, in accordance with our \href{https://www.nexa4ai.com/licenses}{license agreement}.

\section{Future work}
In our forthcoming research, we aim to develop a training framework that accommodates additional data modalities such as audio and video. In addition, we have found that vision input may introduce considerable latency, and we are currently optimizing the inference speed.


\section*{Appendix}
\subsection*{A.1 Selected functions for demo}

For the demonstration, we select 10 representative functions defined on a smartphone, which are described below:

\begin{lstlisting}[style=pythonstyle]
def send_text_message(message_content, phone_number):
    """
    Send the information of the photo as a text message to a specified phone number after understanding the photo.

    Parameters:
    - message_content (str): The content of the text message. This should be a concise
      description or information derived from the photo.
    - phone_number (str): The recipient's phone number in a format recognized by the
      messaging system (usually in international format).

    Returns:
    - bool: Returns True if the message was successfully sent, otherwise False.
    """


def send_email(recipient_email, subject, message_body):
    """
    Send the information of the photo as an email to a specified email recipient after understanding the photo.

    Parameters:
    - recipient_email (str): Only one email address of the recipient.
    - subject (str): The subject line of the email, should be concise.
    - message_body (str): The main content of the email.

    Returns:
    - bool: True if the email was successfully sent, False otherwise.
    """


def google_search(search_query):
    """
    Performs a Google search and retrieves search results based on understanding a photo.

    Parameters:
    - search_query (str): The query string to search for.

    Returns:
    - str: Google search result.
    """


def amazon_purchase(product_description, product_category):
    """
    Initiates a purchase on Amazon for a product based on its description and category, which are derived from analyzing a photo of the product.

    Parameters:
    - product_description (str): A concise description of the product to be purchased, as interpreted from the photo.
    - product_category (str): The category under which the product is listed on Amazon.

    Returns:
    - str: A message detailing the purchase outcome, including the success status, order ID, and total price if successful, or an error message if the purchase fails.
    """


def smart_recycle(item_name, item_category):
    """
    Provide recycling instructions for a specified item based on its category after understanding a photo of the item.

    Parameters:
    - item_name (str): The name of the item to be recycled.
    - item_category (str): The category of the item, must choose one from "electronics", "plastic", "paper".

    Returns:
    - str: Detailed instructions on how to recycle the item properly.
    """


def lost_and_found(item_name, message_content):
    """
    Report a lost item and search for it based on understanding a photo of the item.

    Parameters:
    - item_name (str): The name of the lost item.
    - message_content (str): The description of the lost item.

    Returns:
    - str: Instructions on how to proceed with the lost and found process.
    """


def interior_design(room_type, design_request):
    """
    Generates customized interior design suggestions for a specified room type based on a given design request.

    Parameters:
    - room_type (str): The type of room to be designed, such as "living room", "bedroom", or "office".
    - design_request (str): A specific, concise request for the interior design.

    Returns:
    - str: A comprehensive interior design proposal that includes recommendations for furniture placement, color schemes, and decorative accessories, tailored to the specified room type and design request.
    """


def instacart_shopping(product_name, number_of_order):
    """
    Place an order for a specified quantity of a product using the Instacart shopping service, the product information is from the photo.

    Parameters:
    - product_name (str): The name of the product to be ordered, such as "apples", "bananas", or "pasta".
    - number_of_order (int): The quantity of the product to be ordered.

    Returns:
    - str: Confirmation details that include the order status, total cost, and expected delivery time. If the order cannot be placed, an error message will be provided.
    """


def doordash_order(food_description, address):
    """
    Places an order for delivery of a specified food item to a given address using the DoorDash service, the food description is from the photo.

    Parameters:
    - food_description (str): A description of the food item to be ordered, including any specific preparation requests or dietary considerations.
    - address (str): The delivery address where the food should be sent. Example addresses might include "123 Maple St, Springfield, IL" or "456 Elm Ave, Apt 7, Dallas, TX".

    Returns:
    - bool: True if the order was successfully placed, False otherwise.
    """


def animal_care(animal_type, care_type):
    """
    Provides care instructions for a specific type of animal based on the care needed and frequency, derived from understanding a photo of the animal.

    Parameters:
    - animal_type (str): The type of animal, for example, "dog", "cat".
    - care_type (str): The type of care required, must choose from "feeding", "grooming", "exercise".

    Returns:
    - str: Detailed care instructions tailored to the specific animal type, care needed, and frequency.
    """
\end{lstlisting}

\medskip
{\small
\bibliographystyle{plainnat}
\bibliography{citation}
}


\end{CJK*}
\end{document}